%% file: GenSLAM.tex
\documentclass[letterpaper, 10 pt, conference]{ieeeconf}  

\IEEEoverridecommandlockouts                              

\usepackage{graphicx}
\usepackage{amsmath}
\usepackage{amssymb}
\usepackage{wrapfig}
\usepackage{placeins}
\usepackage{hyperref}

\usepackage{tabu}
\usepackage{subfig}
\usepackage{booktabs}

\usepackage{array}
\newcolumntype{P}[1]{>{\centering\arraybackslash}p{#1}}

\title{\LARGE \bf
GEN-SLAM: Generative Modeling for Monocular Simultaneous Localization and Mapping
}

\author{Punarjay Chakravarty$^{1}$, Praveen Narayanan$^{1}$ and Tom Roussel.$^{2}$
\thanks{$^{1}$ The authors are with Ford Greenfield Labs, Palo Alto, USA.
        {\tt\small pchakra5@ford.com},{\tt\small pnaray11@ford.com}}%
\thanks{$^2$ Work done as an intern for Ford PA, affiliated with KULeuven, Belgium.
    {\tt\small tom.roussel@esat.kuleuven.be}}%
}

\begin{document}

\maketitle
\thispagestyle{empty}
\pagestyle{empty}

\begin{abstract}
We present a Deep Learning based system for the twin tasks of localization and obstacle avoidance essential to any mobile robot. Our system learns from conventional geometric SLAM, and outputs, using a single camera, the topological pose of the camera in an environment, and the depth map of obstacles around it. We use a CNN to localize in a topological map, and a conditional VAE to output depth for a camera image, conditional on this topological location estimation. We demonstrate the effectiveness of our monocular localization and depth estimation system on simulated and real datasets.
\end{abstract}

\section{INTRODUCTION}
The determination of one's position in an environment and the location of other
obstacles around the ego-sensor, are two of the primary tasks
for the sensing system of any robot - be it a service robot in a warehouse
or an Autonomous Vehicle (AV) in a smart city. The sensing used for this is most commonly
some kind of depth sensor - stereoscopic camera or a LIDAR sensor. The computational
and dollar costs of operating these sensors are high. LIDAR sensors still cost in the multiple thousands of dollars, depth cameras don't work well in outdoor lighting conditions and depth obtained from stereo is often noisy. In this paper, we present a method that uses monocular sensing for localization and depth map estimation.
A depth sensor is used to train a network for a particular environment, after
which monocular vision is used to obtain the camera's position in that environment and 
determine a depth-map of obstacles around it.

We use a generative model to learn the joint distribution $p(x_{rgb}, x_{dep})$ between RGB and depth images instead of the traditional discriminative models that estimate $p(x_{dep}|x_{rgb})$.
In what is possibly the first use of generative models for depth estimation in a SLAM context, we use a Variational Autoencoder (VAE) \cite{kingma2013auto} to output
a depth map corresponding to an RGB image. Furthermore, we condition this depth generation on the location of the image in the environment. We obtain this location using 
a standard CNN, trained to output the camera location as one of N topological nodes from the environment.

We are motivated by the underlying assumption that the 3D geometric scene (the depth map) and its 2D projection into the camera (the RGB image) share a common latent subspace. A latent vector sampled from this sub-space, conditioned on location information, should be able to generate both the depth map and its corresponding RGB image for that particular location in the environment.

\begin{figure}[t]
\begin{center}
    \includegraphics[width=0.98\linewidth]{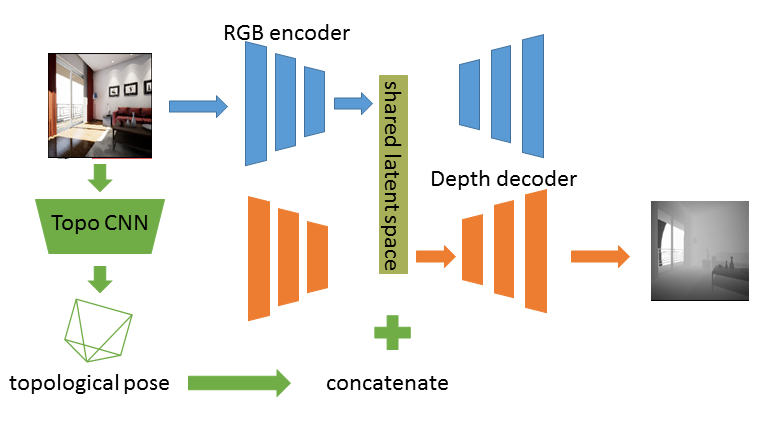}
\end{center}
\vspace*{-0.55cm} 
   \caption{The GEN-SLAM architecture, a conditional VAE with shared latent space in test configuration. The paired RGB and Depth VAEs are indicated in blue and orange respectively. The RGB image is passed through the Topo-CNN to get topological pose, and simultaneously through the RGB encoder to get the latent vector for this image. This shared latent vector is concatenated with the topological pose, and decoded through the Depth decoder, to get the depth map for the RGB image.}
\label{fig:CVAEFwdPass}
\vspace*{-0.95cm} 
\end{figure}

Our Conditional VAE (CVAE) architecture (Figure \ref{fig:CVAEFwdPass}) comprises two encoder-decoder networks, with a shared latent variable, that is conditioned on camera location. The model learns this shared latent representation, with the assumption that latent representations of the RGB image and its paired depth map should coincide. During training, the encoders encode corresponding RGB and depth training samples into a common latent vector, which are then decoded to reconstruct both RGB and depth. Within and cross-domain losses are used to train the network. At test time, the model takes an RGB image, passes it through the RGB encoder, and decodes it using the depth decoder to get the depth map for that image.

Our system trains with a small amount of data. In our experiments, we show that a small number (under 10) laps of a route are enough to train the network to localize in an environment and generate depth maps for each location along its route.

We suggest that this kind of location-aware depth estimation is a useful solution for robots that operate in pre-mapped environments. Topological localization \cite{ulrich2000appearance} is a useful navigation tool, and when used to condition the generation of depth maps from monocular images, it results in better depth maps than without this prior. There are a lot of applications for robots that operate in environments that do not change too much over time. Shelving robots in a warehouse, surveillance robots in a factory, L4 AVs in a geo-fenced urban environment or vacuum-cleaner robots in the home, all operate in areas that can be mapped in advance. This mapping can be done with more expensive sensing -  a ``mapper'' vehicle could have a depth camera or a LIDAR, and cheaper, ``localizer'' vehicles or drones, equipped with monocular vision, can follow pre-mapped routes in that environment and use a system like ours to determine position along that route, as well as get a depth map for local path planning and collision avoidance.

\section{RELATED WORK}
\subsection{SLAM and Localization}
SLAM, initially invented by Leonard et. al. \cite{leonard1991mobile} based on earlier work by Smith et. al. \cite{smith1988estimating}, is the science of estimating one's position in an incrementally built map. Early work was based on single beam LIDAR and sonar, but vision based techniques \cite{newcombe2011dtam,engel2014lsd,mur2017orb} were popularized in the last decade. These papers estimate position in continuous space, which is called metric localization. Topological localization \cite{ulrich2000appearance} refers localization in discrete space, i.e., the localization of a robot closest to a topological node in the environment, and this is done to build more compact maps. When a robot is navigating down a long, featureless corridor for instance, it might not be necessary to know exactly where it is in the corridor, but only that it has reached one of two topological nodes at either end of the corridor. 
In many practical situations, the robot or the sensor operates in previously known environments, and so the mapping and localization aspects of SLAM are separated from each other. However, even if an environment is completely mapped, changes in the environment and dynamic obstacles often mean that the ability to ascertain the distances to unmapped objects in the vicinity of the robot are essential for planning collision-free paths through an environment.

In this paper, we use visual metric localization (ORBSLAM2) \cite{mur2017orb} to train a CNN for topological localization, and also use this topological pose as a conditioning input to our generative model, the CVAE, for the generation of dense depth maps from RGB images.

\subsection{Single image depth}

There has been a glut of work in the past 5 years, that takes a Deep Learning approach to the problem of depth perception from a single image \cite{eigen2014depth,garg2016unsupervised,zhou2017unsupervised,godard2017unsupervised,ummenhofer2017demon,chakravarty2017cnn,li2017undeepvo,ma2017sparse}. 

In the supervised case \cite{eigen2014depth,chakravarty2017cnn}, thousands of image-depth map tuples collected from a depth sensor (like the Microsoft Kinect) are used to train a Convolutional Neural Network (CNN) to determine depth from a single image. However, this requires the use of depth cameras, which is not always available.

More recently, there has been work that uses self-supervision to learn depth in the scene. For example, Garg et. al. \cite{garg2016unsupervised}
uses the  depth map output from the CNN and the left image of a stereo pair to recreate the right image. The reconstructed right image is then compared with the original right image, and this loss is used to train the network. This obviates the need for actual depth maps (which might be noisy), and uses the left and right images from a stereo camera directly. Godard et. al. \cite{godard2017unsupervised} improves upon this work by introducing  left-right consistency in both directions.
SfmLearner, by Zhou et. al. \cite{zhou2017unsupervised} works with monocular sequences for training. It estimates both depth and the camera movement. The depth map and pose output by the network is used to warp an image in a sequence of images coming from a moving camera to its next image in the sequence. The loss between the current image warped to its successor (using the depth map) and the actual successor image is used to train the network.

DeMon \cite{ummenhofer2017demon} and UndeepVO \cite{li2017undeepvo} are recent approaches that also determine both the depth and the motion: rotation/translation matrix $[R \mid t]$ of the monocular camera between frames. Both approaches are also able to determine the absolute scale of the depth map estimated because they use ground truth depth and stereo data respectively.

Mat et. al. \cite{ma2017sparse} uses the sparse depth as would be available from a feature based SLAM system (depth at feature points) to generate a dense depth map. Depth at feature points that would be obtained from a system like ORB-SLAM2 \cite{mur2017orb} are used as an additional input to the CNN, along with the RGB image to generate a dense depth map.

Most of these modern methods use an encoder-decoder architecture for single image depth, and each branch of our twin CVAE uses the same architecture.

\subsection{Deep Generative Models}

Deep generative modeling techniques have shown great promise in many application domains in computer vision \cite{walker-uncertain-future,radford-DCGAN,DRAW,neural-scene-rep,pix2pix,cycleGAN,liu2017unsupervised,huang2018multimodal,sketch-rnn}, natural language processing \cite{generating-sentences-bowman,semeniuta-hybrid-conv-rec,adversarial-feature-matching}, speech \cite{hsu-vcc,hsu-ning-disentanglement,vq-vae} and reinforcement learning \cite{world-models,DARLA}.

Two commonly used classes of Deep Generative models are Generative Adversarial Networks (GANs) and Variational Autoencoders (VAEs) \cite{kingma2013auto, rezende-mohamed-vae}, which are both latent variable models that can construct the data distribution $p(x)$ from a much lower dimensional latent variable distribution $p(z)$. VAEs maximize a variational lower bound for the data distribution, resulting in an encoder-decoder formulation wherein a recognition model (the encoder) maps the data into a lower dimensional latent space, and a reconstruction model or decoder (sometimes called a generator) reconstructs the input, thereby training it as an autoencoder. It differs from an autoencoder in that it one can sample from the latent gaussian prior $z \sim \mathcal{N}(0,I)$, to produce new samples from the data. At test time, the setup becomes a generative model in which data can be produced by sampling from the latent variable distribution $z \sim \mathcal{N}(0,I)$. 

GANs approach the problem of generating samples from the data as a two player adversarial (minmax) game between two networks - a generator $G$ whose aim is to produce data from the data distribution, and a discriminator which classifies data produced by $G$ as real or fake. When the process reaches equilibrium, the generator has learned to fool the discriminator, at which point the samples produced by $G$ mimic those coming from the data distribution. Data samples are produced by sampling a noise variable $\mathcal{N}(0,I)$ and sending it to the generator to obtain $G(z)$. 

GANs and VAEs are generative models with somewhat different origins and motivations. VAEs provide a tractable latent variable formulation, are easy to train, but the samples resulting are known to be somewhat blurry. GANs produce excellent, realistic looking samples but can suffer from mode collapse and training instability. In order to get the best of both worlds, hybrid VAE-GAN models \cite{larsen2015autoencoding} have been proposed so as to obtain a tractable latent variable model with the VAE and use GANs as a polishing mechanism in order to produce good, non-blurry samples. 

A VAE-GAN framework with shared latent space \cite{wan2017crossing,liu2017unsupervised,huang2018multimodal} has recently been used for unsupervised domain adaptation. Images from one domain are transferred into the shared latent space using domain 1's encoder, and decoded into the second domain using domain 2's decoder. Once the domain transfer has been carried out in one direction, it is then carried out in the opposite direction to get back the original image. The difference between the original image in domain 1 its reconstruction, after a cycle to the 2nd domain and back (first introduced by \cite{cycleGAN} as a Cycle Consistency loss), is used to train the network without paired images between domains.

In this paper, we use paired VAEs with shared latent space to generate depth maps from RGB images. We add localization as an added conditioning constraint, and are able to generate images and depth maps for topological nodes in the environment. The following section describes our model in more detail.

\section{GENERATIVE MODELING FOR SLAM}
\subsection{Paired Variational Auto Encoders (VAEs) for RGB and Depth}

\begin{figure}[t]
\begin{center}
    \includegraphics[width=0.98\linewidth]{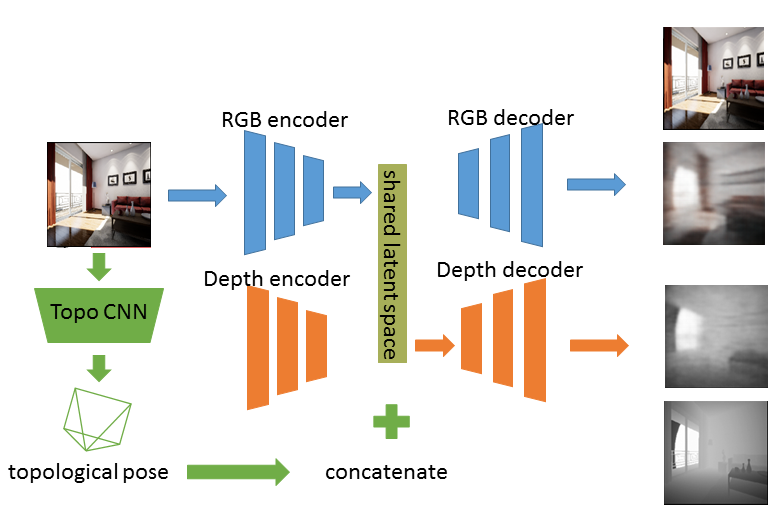}
\end{center}
\vspace*{-0.4cm} 
  \caption{The GEN-SLAM model, with RGB and Depth encoders and decoders, and a shared latent space. It is trained with tuples of rgb, depth and topological pose.}
\label{fig:CVAE_arch}
\vspace*{-0.6cm} 
\end{figure}

Our model has an encoder-decoder architecture with shared latent space for the VAE, as shown in Figure \ref{fig:CVAE_arch}. This paired VAE has 2 encoders and 2 decoders, one each for the RGB and depth images respectively. The encoders encode input RGB images and depth maps into a shared latent space representation, which forces a shared reduced representation of both domains. This latent representation is then decoded through the RGB and depth decoders to give the RGB and depth reconstructions respectively. 

We employ four reconstruction losses to train the network: within-domain reconstruction losses,
$ \mathcal{L}_{x_{rgb} \rightarrow x_{rgb}}$ \& $\mathcal{L}_{x_{dep} \rightarrow x_{dep}}$ and the cross-domain reconstruction losses, $\mathcal{L}_{x_{rgb} \rightarrow x_{dep}}$ \& $\mathcal{L}_{x_{dep} \rightarrow x_{rgb}}$ shown below:

\begin{eqnarray}{}
    & \mathcal{L}_{x_{rgb} \rightarrow x_{rgb}} = \parallel x_{rgb} -  G_{rgb}(E_{rgb}(x_{rgb})) \parallel_{L2} \\
   & \mathcal{L}_{x_{dep} \rightarrow x_{dep}} = \parallel x_{dep} -  G_{dep}(E_{dep}(x_{dep})) \parallel_{L2} \\
   & \mathcal{L}_{x_{rgb} \rightarrow x_{dep}} = \parallel x_{dep} -  G_{dep}(E_{rgb}(x_{rgb})) \parallel_{L2} \\
    & \mathcal{L}_{x_{dep} \rightarrow x_{rgb}} = \parallel x_{rgb} -  G_{rgb}(E_{dep}(x_{dep})) \parallel_{L2} 
\end{eqnarray}

where $x_{rgb}$ and $x_{dep}$ are paired RGB and depth maps used for training and $E$ and $G$ are the Encoders and the decoders (Generators) respectively.

Just using the above losses, we would get a standard AutoEncoder (AE). 
The VAE enforces the latent distribution $q(z)$ to be Gaussian, so that we can sample from it and generate new RGB and depth maps from a trained network. It uses a Kullback-Leibler (KL) distance (and loss) to make the input-conditioned latent distribution $q(z|x)$  to be as close as possible to a 0 mean, unit variance Normal distribution $p_{n}$.

The VAE formulation for the within and cross-domain losses are given as follows:
 
 \begin{align}
 \begin{aligned}
 \mathcal{L}_{x_{rgb} \rightarrow x_{rgb}} = KL(q_{rgb}(z_{rgb}|x_{rgb}) \parallel p_{n}(z)) - \\
 \mathbb{E}_{z_{rgb} \sim q_{rgb} (z_{rgb}|x_{rgb}) } [log \ p_{G_{rgb}}(x_{rgb}|z_{rgb})] 
 \end{aligned}\\
  \begin{aligned}
 \mathcal{L}_{x_{dep} \rightarrow x_{dep}} = KL(q_{dep}(z_{dep}|x_{dep}) \parallel p_{n}(z))   - \\  
 \mathbb{E}_{z_{dep} \sim q_{dep} (z_{dep}|x_{dep}) } [log \ p_{G_{dep}}(x_{dep}|z_{dep})]
 \end{aligned}\\
   \begin{aligned}
 \mathcal{L}_{x_{rgb} \rightarrow x_{dep}} =  KL(q_{rgb}(z_{rgb}|x_{rgb}) \parallel p_{n}(z))   - \\  
 \mathbb{E}_{z_{rgb} \sim q_{rgb} (z_{rgb}|x_{rgb}) } [log \ p_{G_{dep}}(x_{dep}|z_{rgb})]
 \end{aligned}\\
\begin{aligned}
 \mathcal{L}_{x_{dep} \rightarrow x_{rgb}} =  KL(q_{dep}(z_{dep}|x_{dep}) \parallel p_{n}(z))   - \\  
 \mathbb{E}_{z_{dep} \sim q_{dep} (z_{dep}|x_{dep}) } [log \ p_{G_{rgb}}(x_{rgb}|z_{dep})]
 \end{aligned}
 \end{align}
 
where the first terms for each loss value are the KL-divergence terms enforcing the Gaussian constraint on the latent distribution $q(z|x)$ and the second terms are the within and cross domain reconstruction terms. For VAEs, encoding of an input into the latent representation is non-deterministic. For the RGB domain, this is represented by the distribution $ q_{rgb} (z_{rgb}|x_{rgb})$, which can be sampled to give a $z_{rgb}$. This is then passed through the RGB generator $G_{rgb}$ to give a distribution $p_{G_{rgb}}(x_{rgb}|z_{rgb})$. The expected log value of this distribution over multiple samples is the same as the reconstruction RGB reconstruction loss $\mathcal{L}_{x_{rgb} \rightarrow x_{rgb}}$ in equation 1, and the same holds for the other within and cross domain reconstruction losses. For details of this derivation,
see \cite{kingma2013auto}.
Equations 1-4 are used to motivate our discussion on auto-encoders, but we finally use the VAE versions 5-8 to train our networks.
Using these loss functions, we solve the following joint optimization problem in the RGB and depth domains:

\begin{eqnarray}{}
    \min_{E_{rgb}, G_{rgb},E_{dep},G_{dep} } & \mathcal{L}_{x_{rgb} \rightarrow  x_{rgb}} (E_{rgb},G_{rgb}) + \nonumber\\
    &  \mathcal{L}_{x_{dep} \rightarrow x_{dep}}(E_{dep},G_{dep}) + \nonumber\\
    & \mathcal{L}_{x_{rgb} \rightarrow  x_{dep}} (E_{rgb},G_{dep}) + \nonumber\\
     & \mathcal{L}_{x_{dep} \rightarrow  x_{rgb}} (E_{dep},G_{rgb})
\end{eqnarray}

\subsection{Conditional VAE (CVAE)}

The paired VAE for RGB and depth domains is extended with a conditional formulation \cite{walker-uncertain-future} that can be used when the data has additional labels that would be beneficial to the problem. We use the topological node associated with each image as the conditioning input, the assumption being that rgb to depth conversion conditioned on the location information is better than without. This is done in practice by passing the input (RGB or depth) through the encoder to generate the latent code and concatenating the latent code with a one-hot encoding of the topological node. This latent vector, concatenated with location information is then decoded through the rgb and depth decoders as described earlier.  This conditioning also allows us to generate new RGB images and depth maps for the environment, given a particular location. This allows us to hallucinate images for a given topological node, or location in an environment (Figure \ref{fig:topoDreaming}).

\section{IMPLEMENTATION DETAILS}
\subsection{Data Collection}
We generate a synthetic dataset comprising of RGB and depth images using the Unreal gaming engine UE4 \cite{ue4}, and a plugin, UnrealCV \cite{qiu2017unrealcv}. UnrealCV allows us to use OpenCV to programmatically communicate with Unreal Engine and fly the camera through pre-defined trajectories and save RGB images, depth maps and camera poses. We use a living room model from the Unreal Marketplace, and call this the ``Living Room'' dataset.

For real imagery, we desire data where there are not too many dynamic obstacles, and where we can repeat the same route multiple times. We generate our own datasets by using the StereoZed camera \cite{stereozed}, mounting it on a trolley and driving it around the indoor environment of our office. We collect data by doing loops around a lab and a longer loop around the office, and call these the ``Lab'' and ``Corridor'' environments respectively.

\subsection{Topological map Generation}

We use ORB-SLAM2 \cite{mur2017orb} as the conventional SLAM algorithm that supervises the training of our Deep networks. ORB-SLAM2 operates on the stereo image pairs generated by the StereoZed camera and outputs a camera trajectory. To ensure route repeatability, we had to save maps generated by ORBSLAM2 and re-load them for localization for another sequence captured at a different time, along the same route. The original ORBSLAM2 code \cite{orbslam2code} does not come with the ability to save maps. We extended it to allow serializing the map to disk, followed by map re-load and localizing a new sequence of images in the previously saved map. A topological node is set to be at every $1.5$m along the metric route measured by ORB-SLAM2.

\subsection{Network details for the CVAE-model}

\begin{figure}[t]
\begin{center}
    \includegraphics[width=0.99\linewidth]{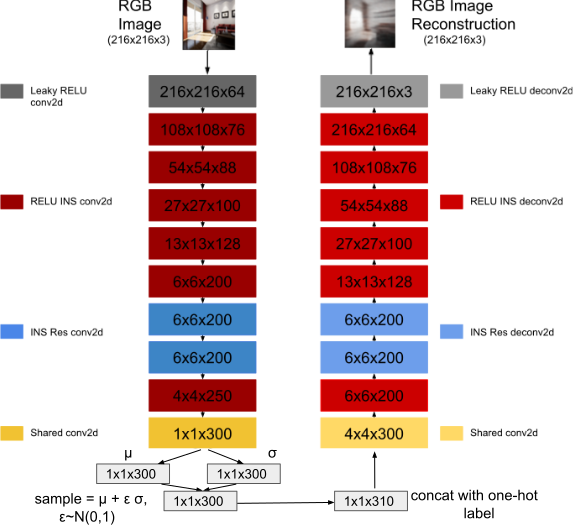}
\end{center}
\vspace*{-0.4cm} 
   \caption{Output tensor sizes for each layer of the RGB VAE. Layers are coloured according to type, with encoder layers (first column) being a darker shade compared to the corresponding decoder layers (second column). The Depth VAE (not shown here) is identical and shares the yellow convolutional layers with the RGB VAE. The topological node in the map the image belongs to is concatenated as a one-hot vector with the sample from the encoder.}
\label{fig:CVAELayers}
\vspace*{-0.3cm} 
\end{figure}

\begin{figure}[t]
\begin{center}
    \includegraphics[width=0.80\linewidth]{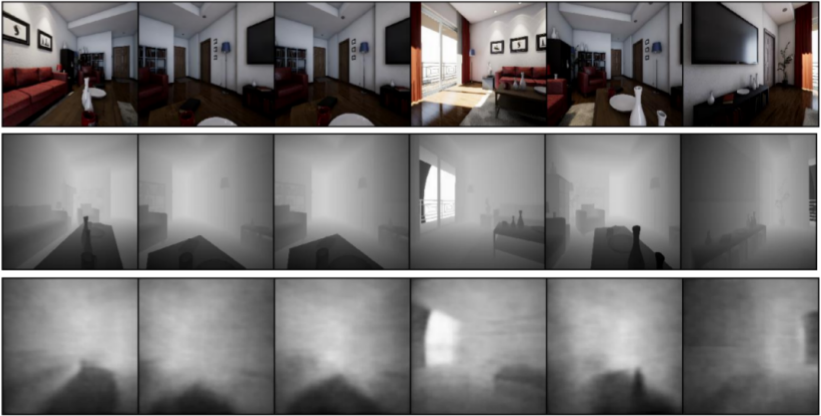}
\end{center}
\vspace*{-0.4cm} 
   \caption{Ground truth (middle row) and estimated (bottom row) depth maps for RGB images (top row) from the Living Room dataset.}
\label{fig:CVAE}
\vspace*{-0.6cm} 
\end{figure}
We use the PyTorch framework for coding our CVAE, which is made location-aware by concatenating the one-hot encoded topological node associated with each image to the latent vector for that image.
We use an additional CNN that operates on the input RGB image, and outputs the topological node as a classification output. We use Alex-net, pre-trained on ImageNet, downloaded from the pytorch model zoo. We remove the last classifier layer, and substitute it with our own, with the number of outputs equal to the number of topological nodes. This pre-trained network is then fine-tuned with (RGB image, topological node) pairs. 
Our CVAE uses convolutional and de-convolutional layers (Figure \ref{fig:CVAELayers}) with instance norm (INS), leaky RELU and residual (Res) layer modifications.
Training of the CVAE is done with (RGB image, depth image, topological node) tuples.

\section{EXPERIMENTAL RESULTS}
We test our proposed method on the data we collected as described previously. For the simulated ``Living Room'' dataset, we generate data by letting the camera travel in trajectories with increments of 0.5 metre offsets + random noise on either side of the reference camera trajectory. For the ``Lab'' and ``Corridor'' datasets we collect data by driving laps of the same loop (12 laps for Lab and 7 laps for Corridor), but with the environment slightly altered (e.g. moving chairs) at different times of the day. We divide these loops into train and test sets using a 90/10 split divided evenly across topological nodes.

Since we use stereo to generate depth maps for Lab and Corridor, there are holes in them, which can negatively impact the training procedure. To address this we use a hole filling algorithm \cite{levin2004colorization} to create dense depth maps. 

\begin{figure}[htb]
\begin{center}
    \includegraphics[width=0.65\linewidth]{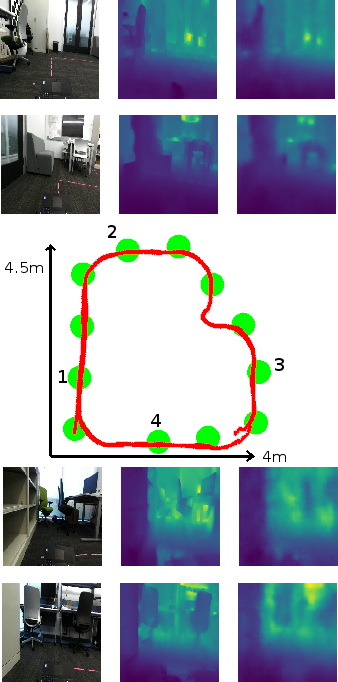}
\end{center}
   \caption{Topological localization and depth map estimation in the Lab dataset. Along rows: rgb, ground truth and estimated depths for 4 topological nodes in map. Metric localization path (red) is obtained from ORBSLAM2.}
\label{fig:topoLocB234}
\vspace*{-0.6cm} 
\end{figure}

\begin{figure}[hbt]
\begin{center}
    \includegraphics[width=0.98\linewidth]{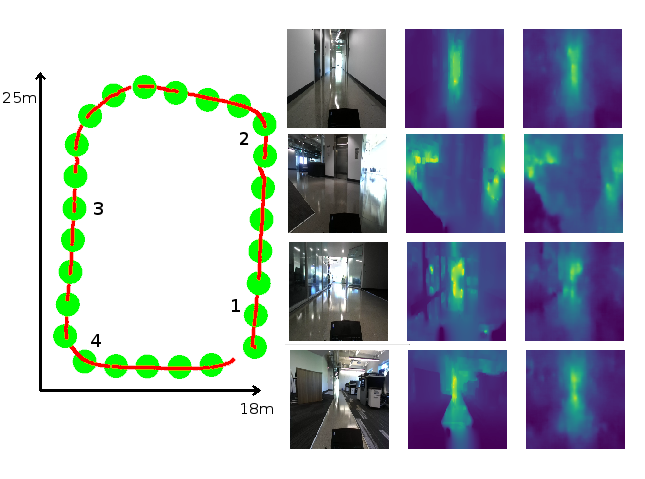}
\end{center}
\vspace*{-0.4cm} 
   \caption{Topological localization and depth map estimation at 4 nodes in the Corridor dataset.}
\label{fig:topoLocCorridor}
\vspace*{-0.3cm} 
\end{figure}

Figures \ref{fig:topoLocB234} and \ref{fig:topoLocCorridor} show topological localization and depth map generation for Lab and Corridor. Left to right along each of the 4 rows: input RGB image, ground truth depth and estimated depth. Centre: topological nodes in green, metric location (for illustrative purposes, from ORBSLAM2) in red and numbers next to nodes correspond to the 4 rows of RGB $\xrightarrow[]{}$  depth reconstruction images.

We show how well our method performs using metrics from ~\cite{garg2016unsupervised, zhou2017unsupervised, godard2017unsupervised}, the current standard for evaluating depth estimation. These metrics are: RMSE, log RMSE, absolute relative difference, squared relative difference and accuracy with different thresholds. 
The absolute relative distance is defined as: $\frac{1}{N}\sum_d \lvert d_{estim} - d_{gt} \rvert / d_{gt}$ and the squared relative difference is defined as: $\frac{1}{N}\sum_d \lVert d_{estim} - d_{gt} \rVert^2 / d_{gt}$.
Accuracy in this context is the ratio of `correct' depth pixels and the total number of depth pixels. A depth pixel is considered correct if: $max(\frac{d_{gt}}{d_{estim}}, \frac{d_{estim}}{d_{gt}}) < \delta$, with $\delta$ being a selected threshold. If the threshold is higher, the metric is less strict.


We show how well our Topological Pose CNN performs using 2 metrics: accuracy and off-by-one accuracy. Accuracy is simply the classification accuracy of our method. The off-by-one accuracy counts the assignment of an image to a node that is adjacent to its actual node as correct. We do this because from a visual standpoint, there is some ambiguity of where one node stops and the other begins. 

The performance of our depth estimator is shown in quantitatively in Table \ref{tab:depth}. Aside from the metrics we add the mean ground truth depth values for each dataset, to give some context to the scale of the different environments. Corridor has a larger mean depth compared to Lab, and hence will naturally result in larger RMSE values when estimating depth, though the other metrics are less sensitive to this. 
The depth values in our simulated dataset Living Room, are not in meters, which partly the reason both RMSE and log RMSE are significantly lower than the other datasets. This environment is also much less complex than our real environments, resulting in more accurate depth estimates. All our non-simulated metrics are computed in meters.

The topological pose CNN accuracy (Table \ref{tab:topotab}) is nearly 100\% across datasets.

We have a video demonstration of the experiments at https://youtu.be/WGuB1cO0mCY

\input{tables/table_recon_results_combined.tex}

\begin{figure}[hbt]
\begin{center}
    \includegraphics[width=0.68\linewidth]{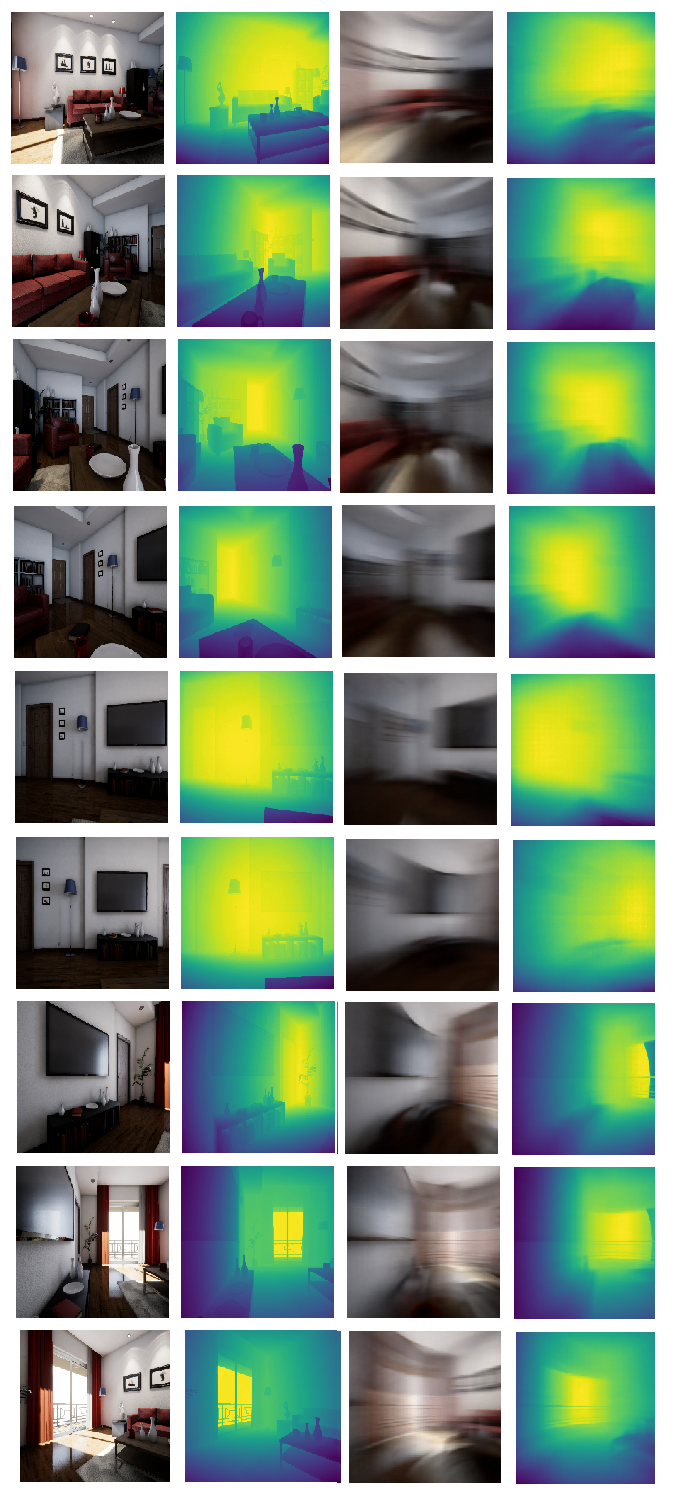}
\end{center}
\vspace*{-0.5cm} 
   \caption{RGB, depth, sampled RGB, sampled depth, for each topological node in the Living Room dataset.}
\label{fig:topoDreaming}
\vspace*{-0.90cm} 
\end{figure}

\section{DISCUSSION AND FUTURE WORK}

One might ask why we use a VAE and not a regular encoder-decoder architecture that has so far been used for single image depth. The VAE forces the latent vector to conform to a distribution, a Gaussian. In our case, both the 3D geometry of the scene (depth map) and its 2D projection (RGB image) conditioned on topological pose, are forced to belong to the same distribution, and the network internalizes the connection between location in an environment, its geometry and appearance, which is embedded in the latent space manifold that the network learns. Figure \ref{fig:topoDreaming} shows the results of sampling RGB images and depth maps from the CVAE for the Living Room dataset. These sampled images are created by sampling z from random noise, concatenating with the topological label, and then passing $[z|label]$ to the RGB and depth decoders. This visualization of sampling from the latent space manifold shows that the network has learnt the appearance and underlying geometry of each topological node.

The twin VAE architecture allows us to use weak supervision to train our networks. Here, we have used noisy depth maps obtained from the stereoZed camera to train our networks. using both within domain and cross domain losses, as opposed to the single cross domain loss that would be used in a regular encoder-decoder architecture that goes from RGB $\xrightarrow[]{}$ depth domains. 

Instead of training using the noisy stereo depth maps, we could use the left and right images of the stereo pair, and train with reconstruction of the right image, given the depth map (produced by the network) \cite{godard2017unsupervised}. We could also train with reconstruction of images over time, as the camera moves through its environment \cite{zhou2017unsupervised}. Cycle Consistency losses \cite{cycleGAN} could also be used to go from RGB $\xrightarrow[]{}$ depth $\xrightarrow[]{}$  RGB, and its converse, depth $\xrightarrow[]{}$ RGB $\xrightarrow[]{}$  depth. These additional loss functions will be examined in future work and should produce better training supervision and allow us to generate better reconstructions of depth in the absence of good training data. 

The topological pose conditioning in this paper could be augmented with additional priors like object detection and lighting for better depth estimation across dynamic environments and different lighting conditions.

Here, our generation of topological maps is relatively simplistic, and operates on single-looped paths. Future work will examine topological map generation for more complicated routes and multi-edged junctions.

\input{tables/table_topo.tex}

\section{CONCLUSIONS}

Mobile robots need to solve the twin tasks of localization and estimating the distance to various obstacles around them to plan collision free paths through an environment. We present a Deep Learning based model that solves both tasks using only monocular vision.
Our Deep Generative model is trained using conventional visual SLAM, and at test time, takes in an RGB image, outputs its topological pose and also generates the depth map for that image, conditioned on that pose. Such a system could be used in an environment where there is a requirement for the operation of multiple robots with cheap sensing. A factory environment, say a fulfillment centre for an e-commerce giant, with hundreds of thousands of square metres of shelving and inventory, and robots to move boxes around, could be mapped once, using depth sensing and conventional geometric SLAM, and this map could be used to train our model. Subsequently, hundreds of robots with single low-cost cameras as sensors could use such a trained model to sense and navigate their way around the factory.

\FloatBarrier

\addtolength{\textheight}{-4cm}   





\bibliographystyle{IEEEtran}


\bibliography{GenSLAM}

\end{document}

%% file: tables/table_recon_results_combined.tex
\begin{table*}
\centering
\caption{Depth estimation metrics across datasets: Living Room, Lab and Corridor.}
\vspace*{-0.3cm} 
\label{tab:depth}
\begin{tabular}{@{}ccccccccc@{}}
\toprule
 & \multicolumn{1}{l}{} & \multicolumn{4}{c}{Lower is better} & \multicolumn{3}{c}{Higher is better} \\ \cmidrule(lr){3-6} \cmidrule(lr){7-9} 
&   &          & &           &          & \multicolumn{3}{c}{Accuracies}                                                                           \\
Dataset & Mean depth  & RMSE     & log RMSE & Abs. Rel. & Sq. Rel. & $\delta < 1.25$ & $\delta < 1.25^2$ & $\delta < {1.25}^3$ \\ \midrule
\textit{Living room}  & 0.69 & 0.031 &	0.083 &	0.039 &	0.002 &	0.968 &	0.994 &	0.998 \\

\textit{Lab [m]} & 1.85 & 0.473 &	0.159 &	0.091 &	0.067 &	0.922 &	0.980 &	0.992 \\
\textit{Corridor [m]} & 4 & 0.908 &	0.211 &	0.140 &	0.183 &	0.844 &	0.956 &	0.980 \\ \bottomrule

\vspace*{-0.8cm}

\end{tabular}
\end{table*}

%% file: tables/table_topo.tex
\begin{table}
\centering
\caption{Topological CNN localization accuracy across datasets}
\vspace*{-0.5cm} 
\label{tab:topotab}
\begin{tabular}{@{}lcc@{}}

\toprule
Dataset              & Accuracy & Off-by-one accuracy \\ \midrule
\textit{Living Room} & 94.35\%      & 100\%                 \\
\textit{Lab}         & 85.77\%  & 99.72\%             \\
\textit{Corridor}    & 97\%     & 100\%               \\ \bottomrule
\vspace*{-0.99cm} 
\end{tabular}
\end{table}